\documentclass[a4paper]{article}
\usepackage{booktabs}
\usepackage{verbatim}
\usepackage{float}
\usepackage{makecell}
\usepackage{color}
\usepackage{ulem}
\usepackage{INTERSPEECH2021}
\usepackage{listings}
\usepackage{caption}
\usepackage{graphicx, subfig}
\usepackage[colorlinks,
            linkcolor=blue,
            anchorcolor=blue,
            citecolor=blue]{hyperref}

\title{Dynamic Multi-scale Convolution for Dialect Identification}
\name{Tianlong Kong$^{\rm{1}}$, Shouyi Yin$^{{\rm{1}}\dag}$\thanks{$\dag$ Corresponding author}, Dawei Zhang$^{\rm{2}}$, Wang Geng$^{\rm{2}}$, Xin Wang$^{\rm{2}}$, Dandan Song$^{\rm{1}}$, \\Jinwen Huang$^{\rm{2}}$, Huiyu Shi$^{\rm{1}}$, Xiaorui Wang$^{\rm{2}}$}
\address{
  $^1$Institute of Microelectronics, Tsinghua University\\
  $^2$Kwai, Beijing, P.R. China}
\email{yinsy@tsinghua.edu.cn}

\begin{document}
\maketitle
	\begin{abstract}

		Time Delay Neural Networks (TDNN)-based methods are widely used in dialect identification. However, in previous work with TDNN application, subtle variant is being neglected in different feature scales. To address this issue, we propose a new architecture, named dynamic multi-scale convolution, which consists of dynamic kernel convolution, local multi-scale learning, and global multi-scale pooling. 
		Dynamic kernel convolution captures features between short-term and long-term context adaptively. Local multi-scale learning, which represents multi-scale features at a granular level, is able to increase the range of receptive fields for convolution operation. Besides, global multi-scale pooling is applied to aggregate features from different bottleneck layers in order to collect information from multiple aspects. The proposed architecture significantly outperforms state-of-the-art system on the AP20-OLR-dialect-task of oriental language recognition (OLR) challenge 2020, with the best average cost performance ($C_{avg}$) of 0.067 and the best equal error rate (EER) of 6.52\%. Compared with the known best results, our method achieves 9\% of $C_{avg}$ and 45\% of EER relative improvement, respectively. 
		Furthermore, the parameters of proposed model are 91\% fewer than the best known model.
		%To be point out, the proposed model owns few parameters of 0.088 times than the best performance architecture published in AP20-OLR-dialect-task.

	\end{abstract}
	\noindent\textbf{Index Terms}: dialect identification, dynamic kernel convolution, local multi-scale learning, global multi-scale pooling.
	
	\section{Introduction}

	Dialect identification refers to the identification of dialect categories from utterances, which is usually presented at the front-end of speech processing systems, such as automatic speech recognition (ASR), multilingual translation systems, targeted advertising, and biometric authentication \cite{li2013spoken,biadsy2011automatic,waibel2000multilinguality,patrick2012language,poorjam2016incorporating}. In recent years, %we have witnessed an increasing interest in spoken or written dialect
    %identification \cite{wang2016ap16,tang2017ap17,ap18,tang2019ap19,li2020ap20}, proven by a high number of evaluation campaigns with more and more participants. 
     due to the raising amount of dialect relating campaigns and participants \cite{wang2016ap16,tang2017ap17,ap18,tang2019ap19,li2020ap20}, the increasing interest in spoken or written dialect identification among them is witnessable.

	Over the past four years, x-vector \cite{snyder2018x} is still the mainstream method for dialect identification.
	Recently, with the flourishing of DNN model, significant architecture improvements in the frame-level layers are springing up, including TDNN \cite{peddinti2015time,snyder2015time}, extended TDNN (E-TDNN) \cite{snyder2019speaker}, factorized TDNN (F-TDNN) \cite{povey2018semi}. ResNet \cite{he2016deep} shows exciting performance in speaker verification and language identification. 
	Inspired by the dense connection in DenseNet \cite{zhu2017densenet}, D-TDNN \cite{yu2020densely} network is adopted to improve accuracy in the field of speaker verification. Compared with the above TDNN-based networks, D-TDNN reduce the number of parameters and further improve the accuracy in speaker verification significantly. 

	However, the difference in dialects is very subtle that distinguishing features of dialects could be local or global in one utterance. Therefore, it is necessary to construct a multi-scale neural network to extract the distinguishing features for dialects.

	In this paper, we propose a novel model for dialect identification, named dynamic multi-scale convolution, for the purpose of further improving the performance and reducing the number of parameters by leveraging dynamic kernel convolution, local multi-scale learning and global multi-scale pooling method. By introducing the methods of dynamic kernel convolution, the features are captured between short-term and long-term context adaptively. Specifically, local multi-scale learning represents multi-scale features at a granular level, which is applied to increase the range of receptive fields for convolution operation. In addition, the reduction of convolutional filter numbers contributes to a great decrement in model parameters. Besides, global multi-scale pooling is applied to aggregate features from different bottleneck layers for collecting information from multiple aspects.

	The contributions of our work are summarized as follows:

\begin{itemize}
\vspace{-0.08cm}
    \item[$\bullet$] Introducing dynamic kernel convolution into dialect recognition for the first time.
    \item[$\bullet$] Local multi-scale learning, which represents multi-scale features at a granular level. 
    \item[$\bullet$] Global multi-scale pooling, which aggregate features from multiple aspects.
\vspace{-0.08cm}
 
\end{itemize}
	
	The rest of this paper is organized as follows. In Section 2 the related works are reviewed, and we describe the dynamic multi-scale convolution architecture in Section 3. The experimental settings are presented in Section 4, and the experimental results are shown and analysed in Section 5. Finally, Section 6 concludes this paper.

	\section{Related Works}
	In this section, we briefly review the related works including the densely connected time delay neural network-statistics-and-selection (D-TDNN-SS) \cite{yu2020densely} and Res2Net \cite{res2net} model. D-TDNN-SS can significantly improve performance of speaker verification in a small number of parameters. 
	%But its performance in the field of dialect recognition is still not optimal. 
	Res2Net is proposed and applied in residual block, improving the performance of ResNet network, significantly.
	%Res2Net is proposed and applied in residual block. It can improve the performance of ResNet network.

	\subsection{DTDNN-SS}

	\cite{yu2020densely} proposed a model named D-TDNN in speaker verification, by adopting bottleneck layers and dense connectivity. D-TDNN comprises fewer parameters than existing TDNN-based models. Furthermore, statistics-and-selection (SS) is proposed in \cite{yu2020densely} to fuse short-term and long-term context from multiple TDNN branches \cite{li2019selective}.

	%Furthermore, they propose an improved variant of D-TDNN, called D-TDNN-SS, to employ multiple TDNN branches with short-term and long-term contexts. D-TDNN-SS can integrate the information from multiple TDNN branches with a newly designed channel-wise selection mechanism called statistics-and-selection (SS).

	%Here, the multi-branch extension of D-TDNN they proposed can adopt multiple TDNN branches with different contexts. 
	%The proposed multi TDNN branches can adopt different frame offsets, so that 
	%In the D-TDNN model, we need to manually set the frame offset of the D-TDNN layer, which may lead to a sub-optimal solution. 
	%The multi-branch extension of D-TDNN can adaptively switch between short-term and long-term context.

	%The SS mechanism is a dynamic channel selection mechanism based on softmax attention. Specifically, the selection module structure in SS is HOSP (High order statistic pooling)-FNN (Feed forward Net)-FNN (Feed forward Net)-Softmax, where HOSP aims to collect channel information from spatial dimensions. The subsequent layers aim to assess the importance of different branches. 

	\subsection{Res2Net}
	Different from the existing methods that represent the multi-scale features in a layer-wise manner, \cite{res2net} propose a novel block for CNNs, named Res2Net, by constructing hierarchical residual-like
	connections within each single residual block. The Res2Net represents multi-scale features at a granular level and increases the range of receptive fields for each network layer.

	\section{Dynamic Multi-scale Convolution for Dialect Identification}
	
\begin{figure}[tb]
		\vspace{-0.2cm}
		\centering
		\includegraphics[scale=0.50]{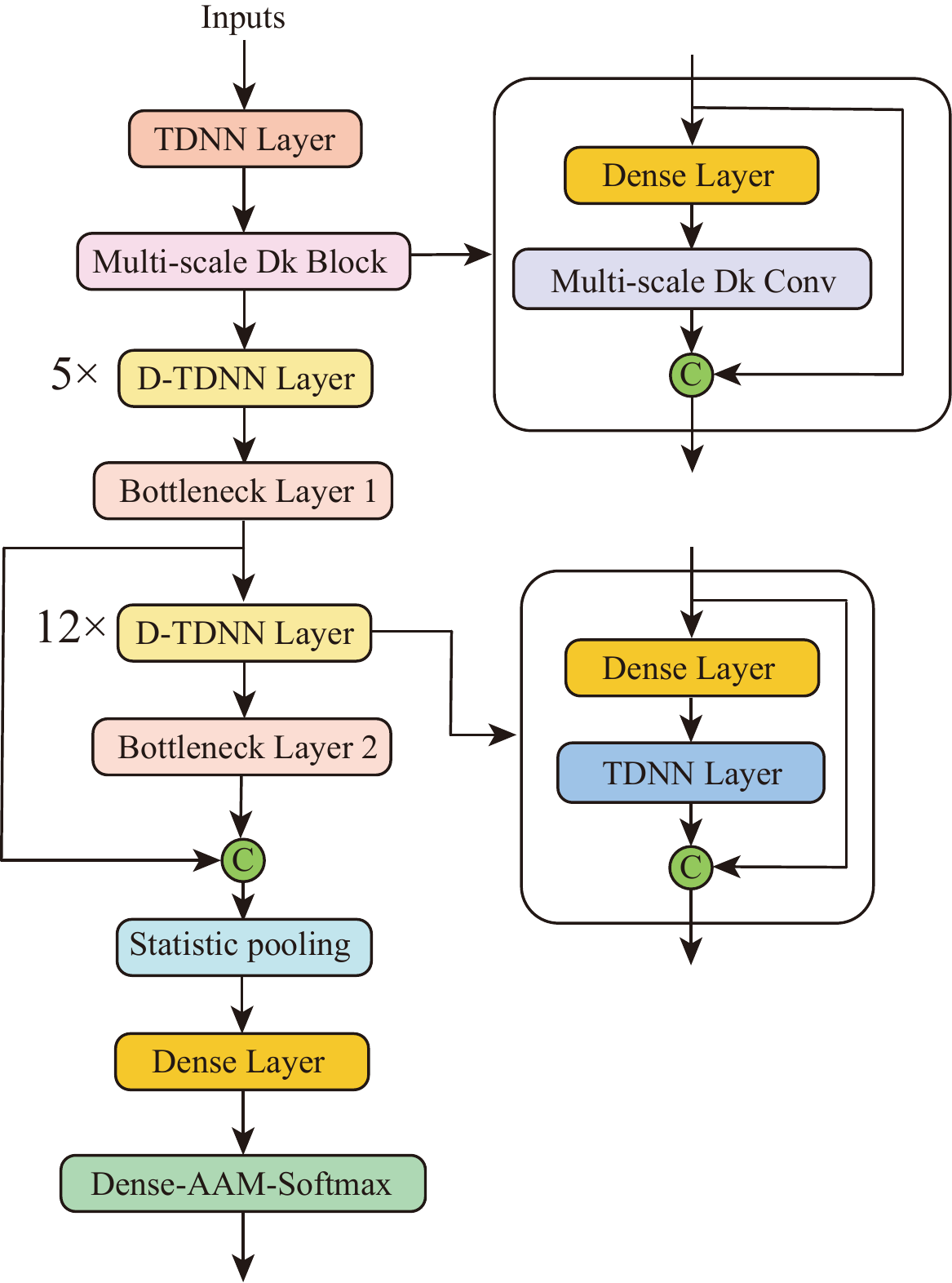}
		\caption{Dynamic multi-scale convolution architecture. In this figure, ``Multi-scale Dk Block'' denotes global and local multi-scale dynamic kernel convolution block, ``Multi-scale Dk Conv'' denotes local multi-scale dynamic kernel convolution operation. Green ``$C$'' icon denotes the operation ``$concat$''.}
		\label{Architecture}
		\vspace{-0.4cm}
	\end{figure}

	In this section, we describe the proposed dynamic multi-scale convolution method for dialect identification. The complete architecture is depicted in Figure \ref{Architecture}. It is noted that batch normalization (BN) and ReLU activation are employed but omitted in the figure.
	%Most existing methods didn't represent the multi-scale features in convolution operation and aggregate different bottleneck layer features to fix more frame-level to utterance-level feature representation.

	The D-TDNN \cite{yu2020densely} network is adopted as the basic skeleton. We modify the first D-TDNN layer to a dynamic multi-scale convolution block, which represents local multi-scale features at a granular level and increases the range of receptive fields for convolution operation. Besides, Global multi-scale pooling aggregates different bottleneck layer features in order to collect information from multiple aspects.

	\begin{figure}[htb]
	\vspace{-0.2cm}
		\centering
		\includegraphics[scale=0.77]{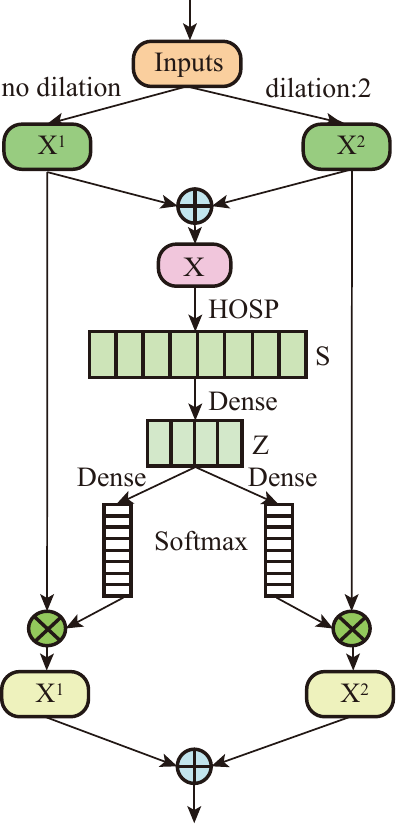}
		\caption{Dynamic kernel convolution (Dk Conv) module. In this figure, $\otimes$ denotes element-wise multiplication. $\oplus$ denotes element-wise addition.}
		\label{SS-Conv}
		\vspace{-0.4cm}
	\end{figure}
	%\vspace{-0.2cm}

	\begin{figure}[tb]
	\vspace{-0.15cm}
	\centering
	\includegraphics[scale=0.77]{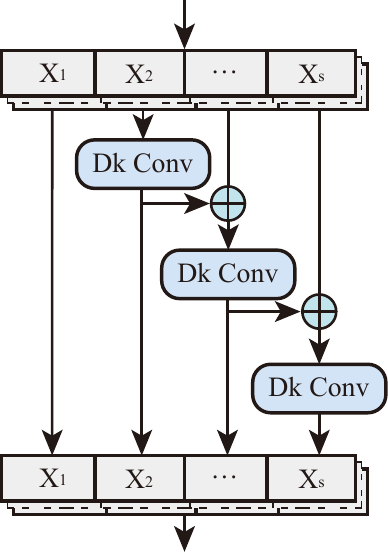}
	\caption{Local multi-scale learning. In this figure, ``Dk Conv'' denotes dynamic kernel convolution operation, $\oplus$ denotes element-wise addition.}
	\label{Conv-SS2}
	\vspace{-0.4cm}
	\end{figure}
	
	\begin{figure}[tb]
	\vspace{-0.2cm}
	\centering
	\includegraphics[scale=0.6]{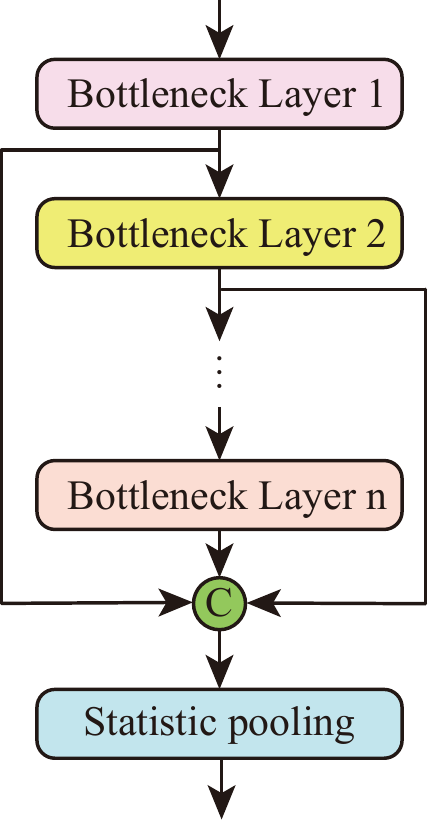}
	\caption{Global multi-scale pooling. In this figure, Green ``$C$'' icon denotes the operation ``$concat$''.}
	\label{channel_aggregated}
	\vspace{-0.4cm}
	\end{figure}

	\subsection{Dynamic Kernel Convolution}

	The dynamic kernel convolution (Dk Conv) is a dynamic channel selection mechanism based on softmax attention. Specifically, the structure of selection mechanism in dynamic kernel convolution is illustrated as followed: high order statistic pooling (HOSP) - dense layer - dense layer - softmax, where HOSP is aimed to collect channel information from spatial dimension. The subsequent layers are aimed to assess the importance of different branches. 

	The dynamic kernel convolution is depicted in Figure \ref{SS-Conv}. Multi-branch extension \cite{zhao2019multi} of convolution captures features between short-term and long-term context adaptively. In proposed model, 
	two branches are used, one of which exploits the methods of no dilation convolution, named $h_d^1$, while the other exploits the methods of the same kernel but dilation convolution, with dilation factor of 2, named $h_d^2$. 
	$h_d^2$ is equivalent to a conventional convolution, which convolutional kernel is different from $h_d^1$.
	First, we combine the information from different convolution by summing up the features extracted from different branches:

	\begin{equation}
	{X = h_d^1 + h_d^2}.
	\end{equation}
	
	Then the HOSP layer collects the mean, standard deviation, skewness and kurtosis information of the sum $X$ for each channel, named ${\mu} \in {R^C}$, ${\sigma} \in {R^C}$, ${s} \in {R^C}$, and ${k} \in {R^C}$.

	%$A = [{a_1}, \cdots ,{a_t}, \cdots ,{a_T}] \in {R^{T \times N}}$

	Given the concatenation of $\mu$, $\sigma$, s, and k, we are able to obtain the softmax-weight after 2 dense layers by the following equation:

	\begin{equation}
	{{s_i}} = \tau {{({W_i}^T({{\rm{V}}^T}[\mu ;\sigma ;s;k] + {b}) + {n_i})}},
	\end{equation}
where ${V}\in {R^{{4C} \times C/r}}$ and ${W_i}\in {R^ {C/r\times {C}}}$ are weight metrics, ${b}\in {R^{{C/r} }}$ and ${n_i}\in {R^{C}}$ are bias items. $\tau$ is the softmax activation function.

	Finally, the dynamic representation ${h_{out}}$ indicates the element-wise dot production sum:

	\begin{equation}
	{h_{out}} = \sum\limits_{i = 1}^2 {{s_i} \otimes h_d^i}.
	\end{equation}

	%\[X = h_d^1 + h_d^2\]
	%DTDNN leverages the idea of DenseNet. As shown in the Figure \ref{Architecture} and \ref{Dense}, the ResNet network obtains the abstract characteristics of Input through short-circuit connections. The final operation of ResNet Block is the element-level addition operation. Compared with ResNet, DenseNet has a denser network connection {}mechanism. The final operation of DenseNet Block is no longer Element-wise addition, but a channel-level splicing operation, so that a more dense network connection can be achieved.

    \subsection{Local Multi-Scale Learning}
    Inspired by the residual connection within ResNet layer in Res2Net \cite{res2net,desplanques2020ecapa}, the local multi-scale learning is adopted to improve representation performance within convolution.
	Local multi-scale learning refers to the multiple available receptive fields at a more granular level. As shown in Figure \ref{Conv-SS2}, we evenly split the feature into $s$ feature subsets, denoted by $X_{i}$, where $i \in [1,2, \cdots, s]$.
	%The number of groups is determined by scale:

	A group of filters firstly extract features from the corresponding feature subsets. Output features of the previous group are then sent to the next group of filters along with another group of input features:
	\begin{equation}
	{\begin{array}{l}
Ou{t_1} = {X_1},\\
Ou{t_2} = F({X_2}),\\
Ou{t_3} = F(Ou{t_2} + {X_3}),\\
          \vdots \\
Ou{t_i} = F(Ou{t_{i - 1}} + {X_i}),\\
          \vdots \\
Ou{t_{s}} = F(Ou{t_{s - 1}} + {X_{s}}),
\end{array}}
	\end{equation} 
	where $F$ denotes the operation of Dk Conv. In Multi-scale Dk Block, the number of Dk Conv filters is divided by $s$.
	The number of operation $F$ filters is the same as D-TDNN layer. After that, we can get the concatenation of $Ou{t_i}$ as the module's output:
	\begin{equation}{
	Out = [{Out_1; } \cdots {; Out_i; } \cdots {; Out_{s}}].
	}
	\end{equation} 

	%This process repeats several times until all input features are processed. 
	Finally, after these feature subsets are processed, features from all groups are concatenated and sent to the next operation to fuse information altogether. By introducing the hyperparameter $s$, local multi-scale learning, which represents multi-scale features at a granular level, is shown effective to increase the range of receptive fields for convolution operation. Besides, with the reduction of convolutional filter numbers, the number of parameters decline significantly.

	\subsection{Global Multi-scale Pooling}
	Previous works \cite{gao2019improving,hajavi2019deep} concluded that feature aggregation at different layers can improve the accuracy of speaker embedding models in speaker verification task. 
	The bottleneck feature is a kind of high-level information integration. Therefore, we aggregate different bottleneck features in channel dimension and send them to statistic pooling layer for the purpose of enhancing the dialect classification capabilities. The structure of global multi-scale pooling method is shown in Figure \ref{channel_aggregated}. 
	%, where two bottleneck layer is adopted for our experiments. 

	We redefine frame-level features ${h_t}$, integrating different bottleneck layer features $h_t^{bi}(i = 1, \cdots ,n)$ in channel dimension, where n is the number of bottleneck layers.

	\begin{equation}{
	{h_t} = [h_t^{b1}, \cdots h_t^{bi}, \cdots ,h_t^{bn}].
	}
	\end{equation} 

	Then global multi-scale pooling layer calculates the mean vector $\mu$
	as well as the second-order statistics in the form of the standard deviation
	vector $\sigma$ over frame-level features ${h_t}(t = 1, \cdots ,T)$.

	\begin{equation}{
	\mu {\rm{ = }}\frac{1}{T}\sum\limits_{t = 1}^T {{h_t}}.
	}
	\end{equation} 

	\begin{equation}{
	\sigma  = \sqrt {\frac{1}{T}\sum\limits_{t = 1}^T {{h_t}}  \odot {h_t} - \mu  \odot \mu } }.
	\end{equation} 

	In experiments, two bottleneck layers are used for global multi-scale pooling.
	Feature representation in terms of global multi-scale pooling, producing higher dialect discriminability to utterance-level features.
	
	\section{Experimental setup}

	\subsection{Datasets}
	In oriental language recognition (OLR) challenge 2020, additional training materials are forbidden to participants, and the permitted resources are several specified data sets \cite{li2020ap20}. We leverage AP17-OL3, AP17-OLR-test, AP18-OLR-test, AP19-OLR-dev, AP19-OLR-test, and AP20-OLR-dialect as our training set for dialect task. All of the training materials are collected from the permitted resources, which include sixteen languages. 
	%We select several utterances of 16 languages from the development set as the validation set and other utterances are used as training set. 
	The detailed information of the combined dataset is shown in Table \ref{Dataset}.
	
\begin{table}[htbp]
	%\vspace{-0.1cm}
	\caption{Dataset for training and evaluation.}
%	\caption{Performance of 2 * 2 convolution kernel. ``-2" means using 2$\times$2 kernels.}
	\label{bookRWCal}
	%\vspace{20pt}
	\centering
	%\resizebox{\textwidth}{!}{{}
	\vspace{-0.1cm}
	\setlength{\tabcolsep}{4.5mm}{
	\begin{tabular}{ccccc}
		\toprule
		\#         & Training     & Validation    & Test \\
	%	\begin{comment}
	%	Model         & \#Param.  & KWS ACC &\makecell[c]{Own/Ex speaker \\ detection \\ ACC} \\
	%	\end{comment}
		\midrule
		
		languages          & 16       & 16   & 6                                    \\
		utterances           & 194,814       & 3168   & 11399                                   \\
		\bottomrule
	\end{tabular}}
	\vspace{-0.3cm}
	\label{Dataset}
\end{table}
	
	\subsection{Implementation Detail}

	\begin{table*}[htbp]
	\vspace{-0.2cm}
	\caption{Comparison for the top system. In this table, Team Royal Flush and Phonexia are the No.2 and No.1 in the leaderboard OLR challenge 2020, respectively. The proposed system achieves 9\% of $C_{avg}$ and 45\% of EER relative improvement, respectively. }
	\label{bookRWCal}
	%\vspace{20pt}
	\centering
	%\resizebox{\textwidth}{!}{{}
	%\vspace{-0.1cm}
	\setlength{\tabcolsep}{4.5mm}{
	\begin{tabular}{ccccc}
		\toprule
		\ Team                     &Parameters (M)      & Model                 & ${C_{avg}}$             & EER(\%) \\
	%	\begin{comment}
	%	Model         & \#Param.  & KWS ACC &\makecell[c]{Own/Ex speaker \\ detection \\ ACC} \\
	%	\end{comment}
		\midrule
		Royal Flush\cite{wangroyal}              &41          & Transformer-12L        & 0.0871                 & 11.97                                   \\
		Phonexia\cite{klcophonexia}                  &33.1         & ResNet18              & 0.0738                 & 11.97                                              \\
		ours                      &2.9    & Global and Local Multi-scale Dk Conv       & 0.0670                 & 6.52                      \\
		\bottomrule
	\end{tabular}}
	\vspace{-0.05cm}
	\label{Compare}
    \end{table*}
	
	Before training, we employ six types of data augmentation, including speed perturbation, the open-source SoX (tempo up, tempo down), the Kaldi recipe \cite{snyder2019speaker}  in bubble, noise, reverb and music disturbance, increasing the amount and diversity of training data. 

	    \begin{table*}[htbp]
	%\vspace{-0.1cm}
	\caption{Results on AP20-OLR-dialect-task.}
%	\caption{Performance of 2 * 2 convolution kernel. ``-2" means using 2$\times$2 kernels.}
	\label{bookRWCal}
	%\vspace{20pt}
	\centering
	%\resizebox{\textwidth}{!}{{}
	%\vspace{-0.1cm}
	\setlength{\tabcolsep}{4.5mm}{
	\begin{tabular}{cccccc}
		\toprule
		\ Model      &Parameters (M)       & Loss Function    & ${C_{avg}}$   &EER(\%) \\
	%	\begin{comment}
	%	Model         & \#Param.  & KWS ACC &\makecell[c]{Own/Ex speaker \\ detection \\ ACC} \\
	%	\end{comment}
		\midrule
		D-TDNN    & 3.3             & Softmax   &    0.0798     &        8.02          \\
		D-TDNN    & 3.3             & AAM-Softmax   &  0.0766       &    7.63             \\
		Dynamic kernel Convolution (Dk Conv)    & 3.4             & AAM-Softmax   &    0.0758    &       7.50          \\
		Local Multi-scale Dk Conv   &2.5              & AAM-Softmax   &    0.0699    &       6.93          \\
		Global and Local Multi-scale Dk Conv  & 2.9              & AAM-Softmax   &   0.0670     &    6.52              \\
		%Aggregated SS-Conv2  & 2.9        & vector-pooling       & AAM-Softmax   &              &                        \\
		\bottomrule
	\end{tabular}}
	\vspace{-0.3cm}
	\label{result}
\end{table*}

	%The 64-dimensional MFCC features are used in our system. Besides,additional 3-dimensional pitch features are appended to acoustic features.

    All features are extracted from 16kHz audio data. 
    %The 67-dimensional splicing feature of 64-dimensional MFCC and 3-dimensional pitch acoustic features are used in our system. 
	The 64-dimensional MFCC features are used in our system. Besides, additional 3-dimensional pitch features are appended to acoustic features.
    All features have frame-lengths of 25ms, frame-shifts of 10ms, and mean normalization over a sliding window of up to 3 seconds. No voice activity detection is applied. The feature engineering is executed using the Kaldi \cite{povey2011kaldi} platform.
    
    Our model is implemented with Tensorflow toolkits \cite{abadi2016tensorflow}.
    At every training step, we sample 16 languages. For each language, a segment with 200 to 400 frames is sliced from the utterances. Additive Angular Margin Softmax (AAM-Softmax) \cite{liu2019large} loss is used to train the baseline system. ${L_{2}}$ regularization is applied to all layers in the network to prevent overfitting. We select stochastic gradient descent (SGD) as the optimizer and the initial learning rate is set to 0.01. 
    %A 3168-utterance validation set is randomly selected from the datasets and the learning rate is halved if the validation loss gets stuck for a while. 
    The loss converges after the learning rate goes down below $10^{-6}$, resulting to around 1.20M training steps. %No dropout is applied in our network following \cite{zeinali2019improve}.

    The D-TDNN layer of our system is the same as the layer presented in \cite{yu2020densely}. For all layers, we select 64 as the number of filters. The hyperparameter $s$ is set to $4$. The context of the first TDNN layer and the last $6$ D-TDNN layers is set to $[t - 5:t + 5]$, and the other layers is $[t - 3:t + 3]$. For dynamic kernel convolution, the kernel is set to $3$.

    \subsection{Evaluation protocol}
    The system is evaluated on the AP20-OLR-dialect-task.  Performance is measured by providing the ${C_{avg}}$, which is defined as the average of the pair-wise performance of test languages, given ${P_{target}}$ = 0.5 as the prior probability of the target language and the Equal Error Rate (EER). A concise ablation study is used to gain a deeper understanding on the effects of each proposed improvement to the overall performance.

	\section{Result and Analysis}

	The dialect identification is evaluated and ranked in OLR challenge 2020. 
	%The principal evaluation metric is ${C_{avg}}$, just as introduced in Section 4.3. And the second evaluation indicator is equal error rate (EER). 
	%We present the dialect identification results of various neural networks by ${C_{avg}}$ and EER. 

	\subsection{Horizontal Result Comparison}
	From Table \ref{Compare}, we can observe that the dynamic multi-scale convolution method performs noticeably better than other teams' model while operating in the same dialect identification task. Compared with OLR Challenge 2020 leaderboard No.1, our model can achieve 9.2\% ${C_{avg}}$ and 45\% EER relative improvement with only 2.9M parameters, respectively.

	\subsection{Result on AP20-OLR-dialect-task}
	Table \ref{result} shows the performance of our ablation study on AP20-OLR-dialect-task. The softmax-output scores of Hokkienese, Sichuanese, and Shanghainese are analysed in the evaluation. 
	All proposed models outperform the state-of-the-art system submitted to OLR challenge 2020 in EER. Noticeably, the proposed dynamic multi-scale convolution method achieves the best performance in all metrics including ${C_{avg}}$, which suggests that the proposed method is effective for dialect identification.

	Results demonstrate that the model with AAM-Softmax is potential to achieve better accuracy than those with softmax loss function. Comparing D-TDNN based method, we discover the operation of dynamic kernel convolution is helpful for dialect identification. Local multi-scale dynamic kernel convolution combines multi-scale learning with dynamic kernel convolution, further improving the performance and reducing 36\% parameters with only 2.5 million by introducing multi-scale learning. Besides, Local multi-scale learning within the convolution is effective in reducing the number of parameters by the hyperparameter $s$. Global and local multi-scale dynamic kernel convolution denotes a variant of local multi-scale DK Conv which adopts global multi-scale pooling. Comparing the results of global and local multi-scale dynamic kernel convolution with local multi-scale Dk Conv result, it can be seen that global multi-scale pooling is excellent to improve the accuracy of dialect identification.

	\section{Conclusions}
	In this paper, we propose a novel dynamic multi-scale convolution model for dialect identification, which introduces dynamic kernel convolution, local multi-scale learning, and global multi-scale pooling. 
	%By introducing dynamic kernel convolution, the network learning can adaptively switch between short-term and long-term context. Local multi-scale learning within convolution can obtain more discriminative information. Besides, global multi-scale pooling can aggragate features of different bottleneck layers so as to collect information from multiple aspects. 

	We evaluate the proposed method with D-TDNN baseline system and carry out the comparison with other submitted systems. Experiments are conducted on OLR challenge AP20-OLR-dialect-task. Significantly, the dynamic multi-scale convolution model achieves the best ${C_{avg}}$ of 0.0670 and EER of 6.52\% in OLR challenge 2020 AP20-OLR-dialect-task. Compared with the known best results, our method achieves 9\% of $C_{avg}$ and 45\% of EER relative improvement, respectively. 
	Furthermore, the parameters of the proposed model are 91\% fewer than the best known model. 
	%In the future, we plan to explore more practive and smaller paramers DNN-based network for dialect identification and evalute the effect of different configurations.

	\section{Acknowledgements}
	This work was supported in part by the China Major S\&T Project (2018ZX01028101-004), the National Key R\&D Project(2018YFB2202600), the NSFC (61774094 and U19B2041)
	and the Beijing S\&T Project (Z191100007519016).
	%The ISCA Board would like to thank the organizing committees of the past INTERSPEECH conferences for their help and for kindly providing the template files. \\
	%Note to authors: Authors should not use logos in acknowledgement section; rather authors should acknowledge corporations by naming them only.
	
	\normalem
	\bibliographystyle{IEEEtran}
	
	\bibliography{mybib}
	
	% \begin{thebibliography}{9}
	% \bibitem[1]{Davis80-COP}
	%   S.\ B.\ Davis and P.\ Mermelstein,
	%   ``Comparison of parametric representation for monosyllabic word recognition in continuously spoken sentences,''
	%   \textit{IEEE Transactions on Acoustics, Speech and Signal Processing}, vol.~28, no.~4, pp.~357--366, 1980.
	% \bibitem[2]{Rabiner89-ATO}
	%   L.\ R.\ Rabiner,
	%   ``A tutorial on hidden Markov models and selected applications in speech recognition,''
	%   \textit{Proceedings of the IEEE}, vol.~77, no.~2, pp.~257-286, 1989.
	% \bibitem[3]{Hastie09-TEO}
	%   T.\ Hastie, R.\ Tibshirani, and J.\ Friedman,
	%   \textit{The Elements of Statistical Learning -- Data Mining, Inference, and Prediction}.
	%   New York: Springer, 2009.
	% \bibitem[4]{YourName17-XXX}
	%   F.\ Lastname1, F.\ Lastname2, and F.\ Lastname3,
	%   ``Title of your INTERSPEECH 2020 publication,''
	%   in \textit{Interspeech 2020 -- 20\textsuperscript{th} Annual Conference of the International Speech Communication Association, September 15-19, Graz, Austria, Proceedings, Proceedings}, 2020, pp.~100--104.
	% \end{thebibliography}
	
\end{document}